\documentclass[11pt]{article}

\usepackage[preprint]{acl}

\usepackage{times}
\usepackage{latexsym}
\usepackage{amsmath}
\usepackage{amssymb}
\usepackage[T1]{fontenc}
\usepackage[most]{tcolorbox}
\usepackage[utf8]{inputenc}

\usepackage{microtype}

\usepackage{inconsolata}
\usepackage{cleveref}

\usepackage{graphicx}

\usepackage{booktabs}
\usepackage{multirow}
\usepackage{colortbl}
\usepackage{xcolor}
\usepackage{pifont}

\definecolor{lightyellow}{rgb}{1.0, 0.98, 0.82}

\newcommand{\codeurl}{\url{https://github.com/copenlu/persona_moe_explain}}

\title{Explaining the Roles of Experts in Mixture-of-Experts LLMs for Personalization}
 \author{
Sekh Mainul Islam,~
Pepa Atanasova,~
Isabelle Augenstein%\vspace{3.5pt}
\\
University of Copenhagen \\
\texttt{\{seis, pepa, augenstein\}@di.ku.dk}
}

\begin{document}
\maketitle
\begin{abstract}
Personalization of Large Language Models (LLMs) is increasingly important for user-centric applications, yet prior methods mainly adapt dense LLMs, whose polysemantic neurons, each encoding multiple unrelated concepts,
hinder localizing the components behind personalization, and thus understanding and controlling it.
% make it difficult to localize important components encoding personalization features, hindering understanding of personalization mechanism and subsequent controlled personalization. 
We propose to personalize Mixture-of-Experts (MoE) LLMs, which sparsely route each token to a few specialized experts, reducing polysemanticity and enabling such localization. We adapt three open-weight MoE LLMs on a large-scale personalization dataset and train crosscoders on paired activations of the personalized and non-personalized models to learn interpretable personalized latents, from which we identify the experts encoding distinct personalization features. The personalized MoE LLMs perform competitively with dense LLMs while activating only a fraction of their parameters, exhibit lower polysemanticity than dense LLMs with similar active parameters, and, within the same model family, yield personalized latents that are less shared with the non-personalized model.
% The personalized MoE LLMs achieve competitive performance than dense LLMs personalization while activating only a fraction of their total parameters, exhibit lower polysemanticity than dense LLMs with similar active parameters, and enable learning personalized latents that are less explained by the non-personalized model than those of dense LLMs.
Causal interventions show that these latents suffice to recover the personalized model's behavior and that boosting their associated experts moves the non-personalized model toward it, revealing distinct routing paths for different personalization features.\footnote{Code and data: \codeurl{}}
% Extensive experiments, including causal analysis, verify that these latents enable stable identification of specialized experts, revealing distinct routing paths for different personalization features.              
\end{abstract}

\begin{figure*}[t]
    \centering
    \includegraphics[width=0.75\linewidth]{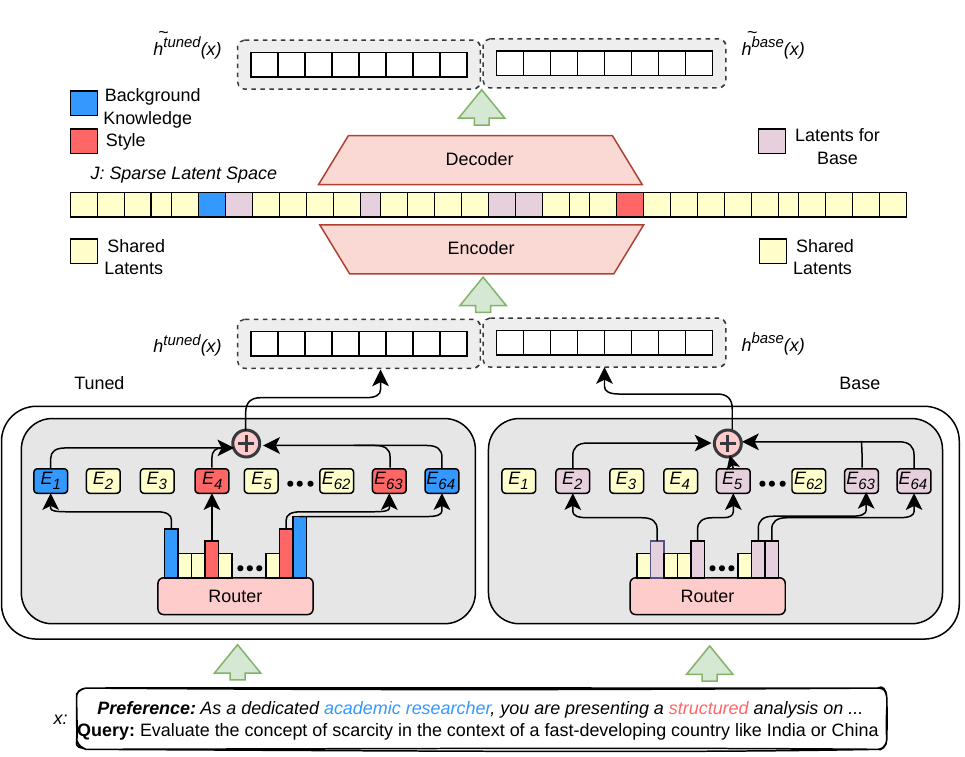}
    \caption{Overview of the proposed framework: [Bottom $\rightarrow$ Top] For a given query and user preference via system message, we collect activations for all tokens of the input prompt from a particular layer of personalized (Left) and non-personalized (Right) MoE and then train a SAE-based model (crosscoder) to learn and identify a set of latents associated with the personalized MoE LLMs capturing distinct personalization features.}
    \label{fig:overall_diagram}
\end{figure*}

\section{Introduction}\label{sec:intro}
Personalization of Large Language Models (LLMs) adapts LLMs to user or group preferences \citep{salemi-etal-2024-lamp,zhang2025personalization}. Due to increasing demand for AI-assisted user-centric applications, Personalization of LLMs has become an important research direction. Most methods use dense LLMs \citep{zhang2025personalization}, offering little interpretability of localizing important model components encoding distinct personalized features due to the polysemanticity of encoding multiple orthogonal concepts into a single neuron \citep{olah2020zoom}. Recent work on developing intrinsically interpretable LLMs suggests that wider, more sparse LLMs enable their neurons to capture more disentangled features and reduce polysemanticity \citep{yang2025mixture}. Mixture-of-Experts (MoE) LLMs offer a scalable solution for wider and sparser networks by replacing a dense FFN block in the MLP layer with a set of experts of the same size, using a sparse routing mechanism to select a subset of these experts to process each input token \citep{shazeer2017}. This mechanism enables MoE LLMs to offer better interpretability for downstream tasks by achieving lower polysemanticity \citep{herbst2026the}, where experts encode distinct task-specific features. Motivated by these observations, we utilize MoE LLMs for personalization and subsequent interpretability analysis, addressing research questions covering polysemamticity of MoE LLMs for personalization (\textbf{RQ1}, \S\ref{sec:rq1}), identifying distinct personalized features encoded via experts during personalization (\textbf{RQ2}, \S\ref{sec:rq2}), and localizing unique routing paths for distinct personalized features (\textbf{RQ3}, \S\ref{sec:rq3}).

Following \citet{bai-etal-2025-understanding} on MoE LLMs alignment, we adapt the MoE LLMs towards personalization using a two-stage parameter-efficient finetuning (PEFT) (\S\ref{sec:method_moe_adaptation}), on a large-scale personalization dataset covering diverse preferences for user groups, MULTIFACETED COLLECTION \citep{lee2024aligning}. We then train a sparse autoencoder-based \citep{bricken2023monosemanticity,huben2024sparse} architecture crosscoder \citep{minder2025overcoming} on paired personalized and non-personalized activations to identify interpretable personalization latents. Mapping these latents to experts reveals distinct routing paths associated with personalization. Our proposed method is illustrated in \Cref{fig:overall_diagram}.     

Extensive experiments on three open-weight instruction-tuned decoder-only MoE LLMs, OLMoE-1B-7B (0125), Qwen3-30B-A3B (2507), and Mixtral-8x7B (v0.1), spanning three personalization benchmarks: MULTIFACETED BENCH \citep{lee2024aligning} covering preferences for user groups, and LaMP and LongLaMP benchmarks for per user preference, verifies the scalability of MoE LLMs in personalization by outperforming similar active parameter sized dense LLMs and achieve competitive performance against similar total parameter sized dense LLMs, while utilizing only $~20\%$ of the total parameters per token (\S\ref{sec:perf_comp}). Furthermore, MoE LLMs enable better learning of crosscoder to identify a high-quality small set of latents (less than $0.25\%$ of total latents), highly contributing in explaining distinct personalized features encoded during personalization than dense LLMs, verifying lower polysemanticity for personalization (\S\ref{sec:rq1}). These personalized latents each capture a distinct personalization feature (\S\ref{sec:rq2}) and map onto a small, stable set of specialized experts (\S\ref{sec:rq3}), revealing distinct routing paths for personalization in MoE LLMs. Causal interventions further confirm that these latents are sufficient to recover personalized behavior and that the identified experts are necessary for it. Overall, our proposed method offers a principled foundation for understanding and designing more transparent personalized language models.

\section{Related Work}
\subsection{Personalization of LLMs}
Methods for aligning LLMs with individual or group preferences \citep{zhang2025personalization} condition on retrieved user history or profile summaries \citep{salemi-etal-2024-lamp,richardson2023integratingsummarizationretrievalenhanced}, encode personas in the prompt \citep{jiang-etal-2024-personallm}, train per-user adapters \citep{tan-etal-2024-democratizing}, learn personalized reward models or merge preference-specific policies \citep{cheng2023deservesrewardlearningcustomized,jang2024personalized}, or steer activations toward a user's style \citep{zhang-etal-2025-personalized,cao2024personalized}. LaMP and LongLaMP \citep{salemi-etal-2024-lamp,kumar2024longlampbenchmarkpersonalizedlongform} evaluate per-user personalization, while Multifaceted Collection and Bench \citep{lee2024aligning} specify group preferences via system messages. These works study dense models as black boxes; we study \emph{how} sparse MoEs represent personalization.\looseness=-1

\subsection{Mixture-of-Experts LLMs.}
MoE layers route each token to a few experts, decoupling parameter count from compute \citep{shazeer2017,JMLR:v23:21-0998,zoph2022stmoedesigningstabletransferable}, and now underlie strong open models such as Mixtral, OLMoE, and Qwen3 \citep{jiang2024mixtralexperts,muennighoff2025olmoe,yang2025qwen3technicalreport}. Analyses of routing report specialization at the level of tokens and syntax rather than domains \citep{jiang2024mixtralexperts,xue2024openmoe}, semantic routing in controlled settings \citep{olson-etal-2025-probing}, and language-specific experts identified by contrasting routing frequencies \citep{bandarkar2026multilingual}. Expert-level specialization can be exploited for adaptation and control: \citet{bai-etal-2025-understanding} tune routers to identify context-faithful experts, \citet{li2026expert} encourage expert divergence during training, and \citet{fayyaz2026steering} steer behavior by (de)activating experts. We adopt a two-stage router-then-expert adaptation for personalization and relate the resulting routing changes to interpretable persona features.

\paragraph{Interpreting model features and fine-tuning.}
Dense-model neurons are polysemantic \citep{olah2020zoom}; sparse autoencoders recover more monosemantic features \citep{bricken2023monosemanticity,huben2024sparse}, sparse probing measures how few neurons encode a concept \citep{gurnee2023finding}, and wider, sparser architectures such as MoEs reduce polysemanticity \citep{yang2025mixture,herbst2026the}. Crosscoders, trained with BatchTopK \citep{bussmann2024batchtopk} and explained automatically \citep{paulo2025automatically}, isolate latents specific to a fine-tuned model and have been applied to chat tuning \citep{minder2025overcoming} and to MoE--dense comparisons \citep{chaudhari2026sparsecrosscodersdiffingmoes}. We diff a personalized MoE against its base and map the resulting latents onto the experts that route them.

\section{Preliminaries}\label{sec:background}

\subsection{Problem Formulation}\label{sec:problem_formulation}
Given a user query $x$ and a preference $P$, either the user's interaction history $P=\{(x_i,y_i)\}_{i=1}^{|P|}$ or a set of stated preferences $P=\{p_1,\ldots,p_n\}$, a personalized LLM $M_\theta$ generates
\begin{equation}
% \small
    \hat{y}_p = M_\theta\big(x, A_p(x, P)\big),
\end{equation}
where the adapter $A_p$ injects $P$ into generation, e.g., via retrieval or learned parameters, and $\hat{y}_p$ is scored against a gold response $y$. We realize $A_p$ as PEFT parameters $\phi$ of an MoE LLM (\S\ref{sec:method_moe_adaptation}) and decompose what it encodes into a set of interpretable latents $\mathcal{J}_{\mathrm{ft}}$ (\S\ref{sec:crosscoder_training_method}).

\subsection{Mixture-of-Experts LLMs}\label{sec:moe_intro}
An MoE layer replaces the dense FFN with $N_E$ experts and a router $r_{\theta_r}$ that routes each token to its top-$k$ experts \citep{shazeer2017,muennighoff2025olmoe}. For token $x_t$ with input $u_{x_t}^{(l)}$ at layer $l$,
\begin{equation}
% \small
    h_{x_t}^{(l)} = u_{x_t}^{(l)} +
    \sum_{i \in \text{top-}k\left(r_{\theta_r}^{(l)}(u_{x_t}^{(l)})\right)}
    g_{(i,t)}^{(l)}\, E_i^{(l)}\big(u_{x_t}^{(l)}\big),
\end{equation}
where $g_{(i,t)}^{(l)}=\text{softmax}\big(r_{\theta_r,i}^{(l)}(u_{x_t}^{(l)})\big)$ is the gate weight of expert $E_i^{(l)}$.

\subsection{Crosscoder}\label{sec:crosscoders}
A crosscoder learns a single dictionary of $D \gg d$ latents that jointly reconstructs paired activations $h_{x_t}^{(l,\mathrm{ft})}, h_{x_t}^{(l,\mathrm{base})} \in \mathbb{R}^d$ of the personalized and non-personalized models at layer $l$ \citep{minder2025overcoming}:
\begin{equation}
% \small
\begin{aligned}
    a_j(x_t) &= \text{ReLU}\big(\mathbf{e}_j^{\mathrm{ft}} h_{x_t}^{(l,\mathrm{ft})} + \mathbf{e}_j^{\mathrm{base}} h_{x_t}^{(l,\mathrm{base})} + b_j^{\mathrm{enc}}\big),\\
    \tilde{h}_{x_t}^{(l,m)} &= \textstyle\sum_j a_j(x_t)\, \mathbf{d}_j^{m} + \mathbf{b}^{\mathrm{dec},m},
\end{aligned}
\end{equation}
where $\quad m \in \{\mathrm{ft}, \mathrm{base}\}$. Each latent $j$ has a separate decoder vector for each model. A latent whose $\mathbf{d}_j^{\mathrm{ft}}$ dominates $\mathbf{d}_j^{\mathrm{base}}$ contributes mainly to reconstructing the personalized model. We use this property to identify personalized latents (\S\ref{sec:crosscoder_training_method}). We train a BatchTopK crosscoder \citep{bussmann2024batchtopk} and the objective and training details are given in \Cref{sec:crosscoders_detailed}.

\section{Methodology}\label{sec:method}
\subsection{Personalization of MoE LLMs}\label{sec:method_moe_adaptation}
Load balancing during MoE pre-training spreads tokens evenly across experts, which discourages individual experts from specializing in task-specific features. Following \citet{bai-etal-2025-understanding}, we therefore adapt the model in two stages: we first tune the routers to find the experts relevant to personalization, and then tune only those experts. Let $\theta=\{\theta_r,\theta_e,\theta_o\}$ denote the router, expert and remaining parameters, and $\mathcal{D}_p$ the preference dataset.

\paragraph{Router tuning.}
We update only the routers with the language-modeling loss:
\begin{equation}
% \small
    \min_{\theta_r}\, \mathcal{L}_{\mathrm{LM}}(M_\theta, \mathcal{D}_p) \quad \text{s.t. } \theta_e, \theta_o \text{ frozen}.
\end{equation}
We then measure how often the tuned router selects each expert $E_i^{(l)}$:
\begin{equation}\label{eq:exp_freq}
% \small
    f_i^{(l)} = \frac{1}{|\mathcal{D}_p|} \sum_{m=1}^{|\mathcal{D}_p|} \frac{1}{|x_m|} \sum_{t=1}^{|x_m|} \frac{\mathbb{I}\big(i \in \mathcal{T}_{m,t}^{(l)}\big)}{k},
\end{equation}
where $\mathcal{T}_{m,t}^{(l)}$ is the set of top-$k$ experts selected for token $t$ of sequence $x_m$ at layer $l$. Dividing by $k$ ensures that the frequencies at each layer sum to 1.

\paragraph{Expert tuning.}
We then fine-tune only the $k$ most selected experts per layer, $\theta_e^{(k)} \subset \theta_e$:
\begin{equation}
% \small
    \min_{\theta_e^{(k)}}\, \mathcal{L}_{\mathrm{LM}}(M_\theta, \mathcal{D}_p) \quad \text{s.t. } \theta_r, \theta_o \text{ frozen}.
\end{equation}

\subsection{Disentangling Personalized Features}\label{sec:crosscoder_training_method}
We train a crosscoder (\S\ref{sec:crosscoders}) on paired activations of the personalized and non-personalized MoE LLMs over $\mathcal{D}_p$, identify the latents specific to the personalized model, and map each latent to the experts that process it.

\textbf{Layer selection.}
Middle layers carry the most informative representations \citep{skean2025layer}.%minder2025overcoming 
Among them, we select the layer whose routing changes most under personalization, measured by the Jensen--Shannon divergence between the routing distributions of personalized vs. non-personalized models \citep{li2026expert,bandarkar2026multilingual}. We require the router to choose experts confidently, as measured by low routing entropy.

\textbf{Personalized latents.}
Following \citet{minder2025overcoming}, we score each latent by the relative norm difference of its two decoder vectors,
\begin{equation}
% \small
  \Delta_j = \frac{1}{2}\left(\frac{\lVert \mathbf{d}_j^{\mathrm{ft}} \rVert_2 - \lVert \mathbf{d}_j^{\mathrm{base}} \rVert_2}
  {\max\!\left(\lVert \mathbf{d}_j^{\mathrm{ft}} \rVert_2,\, \lVert \mathbf{d}_j^{\mathrm{base}} \rVert_2\right)} + 1\right) \in [0,1],
\end{equation}
and assign $j$ to $\mathcal{J}_{\mathrm{ft}}$ if $\Delta_j \geq 0.9$, to $\mathcal{J}_{\mathrm{base}}$ if $\Delta_j \leq 0.1$, and to $\mathcal{J}_{\mathrm{shared}}$ if $\Delta_j \in [0.4, 0.6]$.

\textbf{Latent-to-expert mapping.}
For each $j \in \mathcal{J}_{\mathrm{ft}}$, we take the top-$N$ sequences in $\mathcal{D}_p$ by activation $a_j$. Within these sequences, we rank the experts at each layer by how much more often they are selected on tokens where the latent fires than on tokens where they do not \citep{bandarkar2026multilingual}, and keep the top-$k'$:
\begin{equation}
% \small
  \mathcal{E}_j^{(l)} = \mathrm{Top}_{k'}\Big(f_i^{(l)}\big|_{a_j>0} - f_i^{(l)}\big|_{a_j\leq0}\Big),
\end{equation}
where $f_i^{(l)}$ is the expert activation frequency (\Cref{eq:exp_freq}) restricted to the given tokens. We choose $(N, k')$ by grid search over two criteria in $[0,1]$. \textit{Distinctiveness} ($D$) is one minus the average overlap of $\mathcal{E}_j^{(l)}$ across latents, so different latents should map to different experts. \textit{Stability} ($S$) is the agreement of $\mathcal{E}_j^{(l)}$ across random subsamples of the $N$ sequences, so the mapping should not depend on which examples are drawn. Larger $N$ improves stability, but if $N$ is too large, the selected sequences overlap across latents and distinctiveness drops.

\section{Experimental Settings}\label{sec:exp_setting}
\textbf{Data and evaluation.}
We personalize models and train crosscoders on \textsc{Multifaceted Collection} \citep{lee2024aligning}. Each example pairs a user query with a system message that encodes a group-level preference across four dimensions (style, background knowledge, informativeness, harmlessness), along with a preference-aligned response generated by GPT-4-Turbo-0125. We evaluate on three benchmarks:
(i) \textsc{Multifaceted Bench} \citep{lee2024aligning}, which uses the same system-message format but has no preferences in common with the training set;
(ii) LaMP \citep{salemi-etal-2024-lamp}; and
(iii) LongLaMP \citep{kumar2024longlampbenchmarkpersonalizedlongform}.
LaMP and LongLaMP instead require inferring per-user preferences from interaction history. Following \citet{zhang-etal-2025-personalized}, we use their user splits and omit the email tasks, whose data is not publicly available. We report METEOR and ROUGE-L against the gold responses.

\textbf{Models and baselines.}
We study three instruction-tuned MoE LLMs: OLMoE-1B-7B (0125), Qwen3-30B-A3B (2507) and Mixtral-8x7B (v0.1). Each is evaluated without adaptation, after router tuning (RT), and after router and expert tuning (RT+EFT). We compare each MoE model with two dense models, one matching its active and one matching its total parameter count: OLMo-2-1B/7B, Qwen2.5-3B/32B, and Llama-2-13B/70B for Mixtral, following its technical report (\Cref{tab:model_params}). The dense baselines are evaluated without adaptation and after PEFT on \textsc{Multifaceted Collection}. As a reference, we also report \textsc{Janus}-7B and its ORPO/DPO variants \citep{lee2024aligning}, which are fully parameter-fine-tuned on the \textsc{Multifaceted Collection}, indicating the upper bound of performance for LLMs with similar parameter counts.

\begin{figure}[t]
    \centering
    \includegraphics[width=0.9\linewidth]{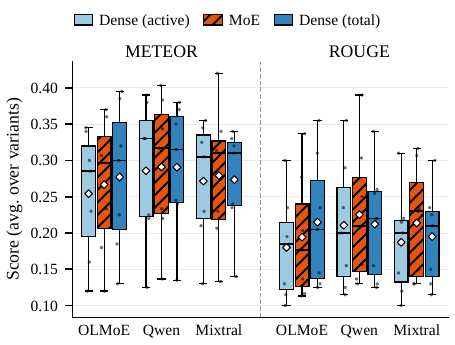}
    \caption{Personalization performance of LLMs: each box plot shows average performance over personalized and non-personalized variants on 7 datasets (points), with the mean performance represented using $\diamond$. The detailed result is described in \Cref{tab:llm_personalization_comp}.}
    \label{fig:personalization_perf_comp}
\end{figure}

\begin{figure}[t]
    \centering
    \includegraphics[width=0.9\linewidth]{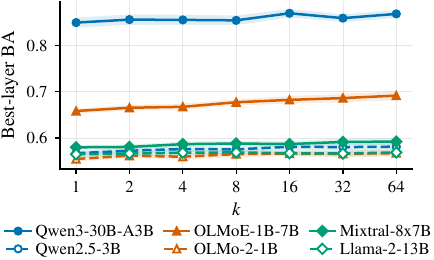}
    \caption{K-sparse probe balanced accuracy comparison between MoE and dense LLMs with similar active parameters.}
    \label{fig:moe_vs_dense_llm_polysemanticity}
\end{figure}

\begin{figure}[t]
    \centering
    \includegraphics[width=0.9\linewidth]{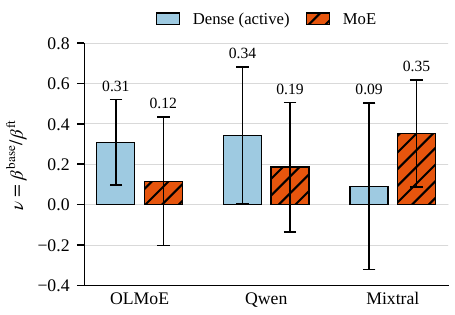}
    \caption{Ratio $\nu$ of scales to explain non-personalized and personalized models via personalized latents and their behavior across dense and MoE LLMs.}
    \label{fig:beta_comp}
\end{figure}

\section{Results}\label{sec:results}
%We now compare the personalization performance of MoE LLMs with dense LLMs and discuss the research questions to uncover the unique routing paths for personalization in MoE LLMs.

\subsection{Personalization performance comparison}\label{sec:perf_comp}
Across model families and evaluation metrics, \textit{MoE LLMs outperform dense LLMs with similar active parameters}, and for the Qwen and Mixtral families, they match or exceed dense LLMs with similar total parameters (\Cref{fig:personalization_perf_comp}). The per-variant results in \Cref{tab:llm_personalization_comp} show the same pattern after personalization: personalized MoE LLMs outperform or match personalized dense LLMs with similar active parameters. Router tuning accounts for most of this gain in Qwen3, whereas expert tuning contributes more in Mixtral. \textit{Across benchmarks, personalization yields the largest gains on MULTIFACETED BENCH}, which shares the training data's (MULTIFACETED COLLECTION) form of preference (group-level preferences given via system messages), and smaller gains on LaMP and LongLaMP, which require transferring to per-user preferences inferred from interaction history. \textit{Overall, MoE LLMs achieve competitive personalization while activating only a fraction of their parameters, motivating a mechanistic analysis of how their experts encode personalization features.}

\subsection{RQ1: Do Sparse MoE LLMs offer lower polysemanticity, enabling better disentanglement of personalized features?}\label{sec:rq1}
\textbf{Polysemanticity.} To investigate the polysemanticity behavior of neurons in MoE experts compared to dense LLMs, we use k-sparse probing \citep{gurnee2023finding}. Following \citet{herbst2026the}, we estimate the mean difference of activations between positive and negative tokens per concept and compute the binary classification of a probe considering only the top-$k$ neurons with the highest activation difference. For our personalization task, we consider the top 5 most frequent sub-dimensions from  MULTIFACETED COLLECTION as concepts for each dimension, resulting in 20 independent concepts. For each concept, we take the response tokens (up to 1024 per response, 100 queries) of samples whose system message specifies that sub-dimension as positives, and those of samples specifying a different sub-dimension of the same dimension as negatives (e.g., \textit{formal} vs.\ \textit{casual} under style), giving about 50{,}000 tokens per class.
% Per concept, we consider all response tokens (up to 1024) corresponding to queries (100) as positive tokens, where the system preference includes that concept (sub-dimension). Response tokens from samples with preferences from different sub-dimensions for the same dimension are considered as negative tokens, resulting in around 50000 tokens per class. 
Following \citet{herbst2026the}, we only consider tokens routed to the target expert for MoE LLMs and identify the best layer containing the expert achieving the highest classification accuracy on the test split (25\% split). For dense LLMs, the best layer is that with the highest-performing MLP. \Cref{fig:moe_vs_dense_llm_polysemanticity} illustrates the variance in k-sparse probe classification accuracy from $k=64$ to $k=1$ for the MoE and dense LLMs with similar active parameters. MoE LLMs outperform dense LLMs with similar active parameters in the balanced accuracy metric, and the performance gap is wider for the same model family (Olmo, Qwen). The gap remains wide at $k=1$, \textit{verifying monosemantic behavior of MoE experts for personalization.}  

\textbf{Disentanglement of personalized features.} To verify whether lower polysemantic MoE experts enable better disentanglement of personalised features via crosscoder latents, we compare MoE and dense LLMs on how well personalized latents explain activations from non-personalized LLMs. In particular, following \citet{minder2025overcoming}, we compute scales $\beta^{\text{base}}_j, \beta^{\text{ft}}_j$ for each personalized latent $j \in \mathcal{J}_{\mathrm{ft}}$ in explaining activations from non-personalized and personalized model as:
\begin{equation}
% \small
    \beta^{\text{base/ft}}_j = \operatorname*{argmin}_{\beta} \sum_{i=1}^{n} \left\| \beta a_j(x_i) d^{\text{ft}}_j - h^{\text{base/ft}}(x_i) \right\|_2^2
\end{equation}

Since, $\beta^{\text{base}}_j, \beta^{\text{ft}}_j$ are unbounded, their ratio $\nu_j = \beta^{\text{base}}_j / \beta^{\text{ft}}_j$ approaching near zero indicate little contribution to reconstructing the non-personalized model relative to the personalized model.
% approaching zero indicates `pure' personalized latent, it approaching 1, indicates these latents are equally shared by both variants and are less effective in disentangling personalized features. 
\Cref{fig:beta_comp} compares the $\nu_j$ corresponding to personalized latents of MoE and dense LLMs. For the same model family, the MoE LLM has a lower $\nu_j$ for personalized latents than dense LLMs, except for Mixtral, which yields a higher $\nu$ than a similar active-parameter-sized Llama-13B model, indicating that this comparison is limited to the same model family. \textit{Overall, MoE LLMs offer lower polysemanticity and better disentanglement of personalized features, which justifies the need for interpretability analysis for decoding personalization mechanism.}     

\begin{figure}[t]
  \centering
  \includegraphics[width=0.9\columnwidth]{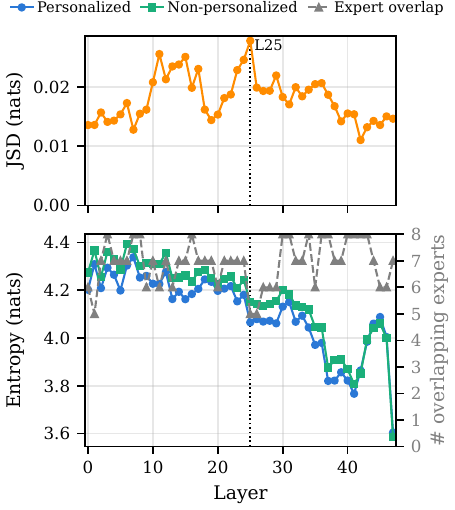}
  \caption{Layer-wise routing distribution analysis. \textbf{Top:} routing JS-divergence between the personalized and non-personalized Qwen3-30B-A3B (2507). \textbf{Bottom:} entropy of the two models with overlapped experts (right axis). L25 is identified as the best layer. Refer to \Cref{fig:layerwise_routing_jsd_entropy_omloe,fig:layerwise_routing_jsd_entropy_mixtral} for other models.}
  \label{fig:layerwise_routing_jsd_entropy_qwen3}
\end{figure}

\begin{figure}[t]
    \centering
    \includegraphics[width=0.9\linewidth]{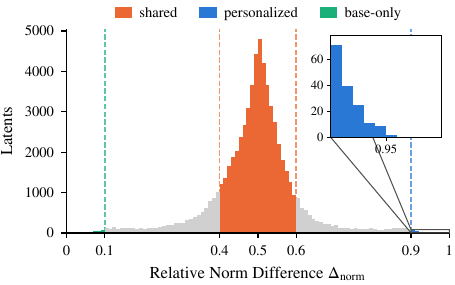}
    \caption{Relative norm difference distribution of latents from the crosscoder trained on activations from L25 of personalized and non-personalized Qwen3-30B-A3B (2507). Refer \Cref{fig:latent_norm_dist_olmoe,fig:latent_norm_dist_mixtral} for other models.}
    \label{fig:latent_norm_dist_qwen}
\end{figure}

\begin{figure}[t]
    \centering
    \includegraphics[width=0.9\linewidth]{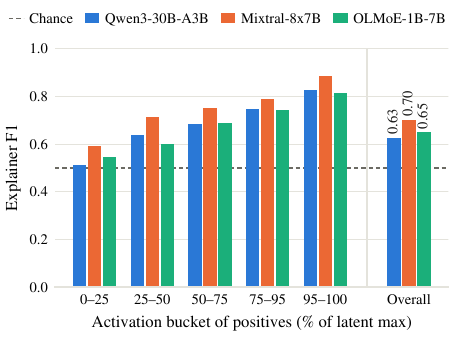}
    \caption{F1 for %of the scorer in
    classifying positive and negative examples activated by latents (Overall). We also show its efficiency across different activation quantiles.}
    \label{fig:explainer_f1}
\end{figure}

% Preamble: \usepackage{xcolor} \usepackage[most]{tcolorbox}
\definecolor{moe}{HTML}{E6550D}
\definecolor{moelight}{HTML}{FDF0EA}
\begin{figure}[t]
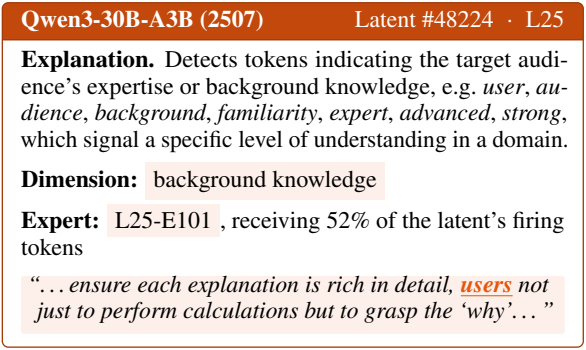

\centering
\begin{tcolorbox}[enhanced, width=\columnwidth, boxrule=0.5pt, arc=2pt,
  colback=white, colframe=moe!70!black, left=4pt, right=4pt, top=1pt, bottom=2pt,
  title={\small\textbf{Qwen3-30B-A3B (2507)} \hfill Latent \#48224 \;$\cdot$\; L25},
  coltitle=white, colbacktitle=moe!85!black, fonttitle=\small]
\small
\textbf{Explanation.} Detects tokens indicating the target audience's expertise or
background knowledge, e.g.\ \textit{user}, \textit{audience}, \textit{background},
\textit{familiarity}, \textit{expert}, \textit{advanced}, \textit{strong}, which signal a specific
level of understanding in a domain.

\smallskip
\textbf{Dimension:} \colorbox{moelight}{background knowledge}\\[0.5pt]
\textbf{Expert:} \colorbox{moelight}{L25-E101}, receiving 52\% of the latent's firing tokens
\begin{tcolorbox}[colback=moelight, colframe=moelight, boxrule=0pt, arc=1pt,
  left=3pt, right=3pt, top=-3pt, bottom=1pt]
\footnotesize\textit{``\ldots ensure each explanation is rich in detail,
\textcolor{moe}{\textbf{\underline{users}}} not just to perform calculations but to grasp
the `why'\ldots''}
\end{tcolorbox}
\end{tcolorbox}
\caption{Top activating latent from Qwen3-30B-A3B (2507): its autointerpretability explanation, dominant preference dimension in MULTIFACETED COLLECTION, the top-$k'$ expert at layer 25 with the share of the latent's activated tokens routed to it, and an excerpt from a top-$N$ activating sequence with the maximally activating token highlighted. More top activating latents from different models are described in \Cref{tab:latent_analysis}.}
\label{fig:qwen_latent_example}
\end{figure}

\subsection{RQ2: What are the different personalized features disentangled into a set of interpretable concepts?}\label{sec:rq2}
% To understand the personalization mechanism of MoE LLMs, we decompose the personalized features encoded via experts in MoE LLMs during personalisation into a set of interpretable concepts (latents) from a crosscoder trained on paired activations from personalized and non-personalized MoE LLMs. To explain the most distinctive and informative personalized features via crosscoder, we collect activations from an optimal layer for the personalized and non-personalized MoE LLMs based on routing divergence and routing entropy, as 
Using the routing-divergence and entropy criteria discussed in \S\ref{sec:crosscoder_training_method}, we identify L25, L15, and L6 as the most distinctive layers for Qwen3-30B-A3B (2507), Mixtral-8x7B (v0.1), and OLMoE-1B-7B (0125) respectively, as illustrated in \Cref{fig:layerwise_routing_jsd_entropy_qwen3,fig:layerwise_routing_jsd_entropy_mixtral,fig:layerwise_routing_jsd_entropy_omloe}. As discussed in \S\ref{sec:crosscoder_training_method}, we identify a very small subset (less than 0.25\%) of latents with very high ($\geq 0.9$) decoder norm to reconstruct activations from the personalized MoE LLMs. In particular, we identify (157/65536), (132/131072), (111/65536) personalized latents for Qwen3-30B-A3B (2507), Mixtral-8x7B (v0.1), and OLMoE-1B-7B (0125) respectively, as illustrated in \Cref{fig:latent_norm_dist_qwen,fig:latent_norm_dist_mixtral,fig:latent_norm_dist_olmoe}.\looseness=-1

\textbf{Interpreting personalized latents.} Following \citet{paulo2025automatically,minder2025overcoming}, we explain each personalized latent with an LLM explainer and evaluate the explanation with an LLM scorer (both Llama-3.3-70B-Instruct). The explainer sees 30 activating examples (the top 20 and 10 weaker ones), each a 40-token window centred on the maximally activating token (32 tokens before, 8 tokens after). The scorer then labels 20 examples as activating and 20 as non-activating, using only the explanation. \Cref{fig:qwen_latent_example} shows an example: a max-activating latent from Qwen3 is explained as tracking the audience's level of expertise.

Most personalized latents encode the \emph{requested} preference rather than the response: 85\% of OLMoE and 82\% of Qwen3 latents fire mainly on the system message. Mixtral is the exception, with 41\% firing mainly in the response. The models also emphasize different preference dimensions: OLMoE's latents are dominated by style (52\%) and harmlessness (21\%), whereas Qwen3's and Mixtral's are dominated by background knowledge (45\% and 40\%). The scorer reaches moderate F1 overall (0.65 OLMoE, 0.63 Qwen3, 0.70 Mixtral; \Cref{fig:explainer_f1}), rising monotonically across activation bins to 0.81, 0.83 and 0.88 in the top bin, so the explanations best describe strong activations. \textit{Overall, the crosscoder decomposes the features that personalization adds to MoE LLMs into a small set of interpretable concepts.}

\begin{figure}[t]
  \centering
  \includegraphics[width=0.9\columnwidth]{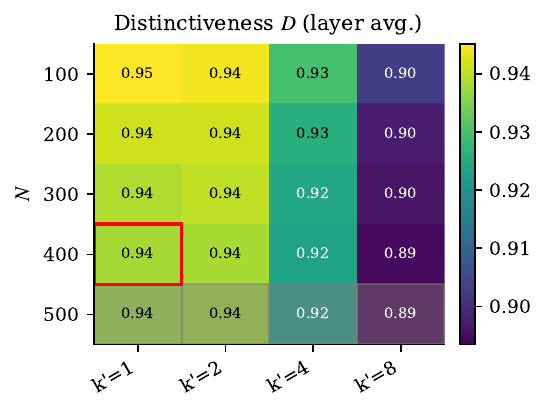}\\
  \includegraphics[width=0.9\columnwidth]{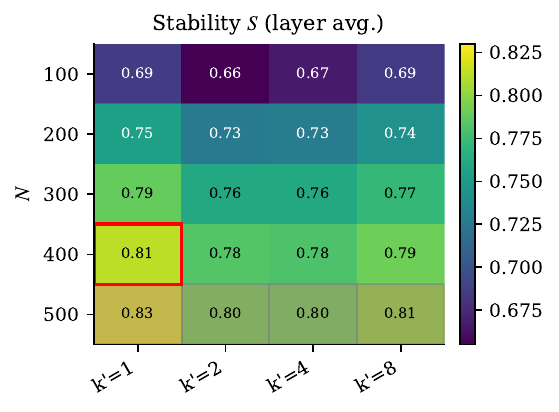}
  \caption{Grid search over $(N,k)$ for Qwen3-30B-A3B (2507). \textbf{Top:} Distinctiveness of top-$k'$ experts across latents with varying $N$. \textbf{Bottom:} Stability of top-$k'$ experts for the same $N$ examples under resampling for a particular latent.}
  \label{fig:latent_to_expert_mapping_hyperparameter_qwen}
\end{figure}

\begin{figure}[t]
  \centering
  \includegraphics[width=0.9\columnwidth]{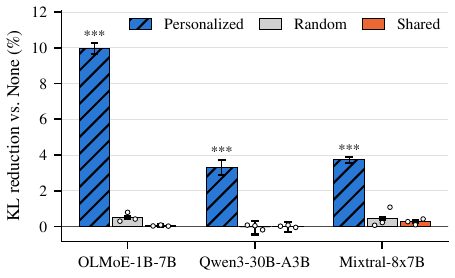}\\[2pt]
  \includegraphics[width=0.9\columnwidth]{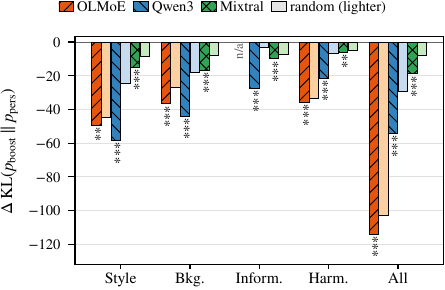}
  \caption{\textbf{Top:} KL patching at the crosscoder layer over full responses.   `Random' and `Shared' baselines are shown as the mean of 3 draws with dots. \textbf{Bottom:} Steering top-$k'$ experts per dimension in the non-personalized model; measuring KL divergence with its personalized counterpart. ***: statistically significant with $p<0.001$ vs.\ controls (Wilcoxon, Holm-corrected).}
  \label{fig:kl_patching_latent_expert_boosting}
\end{figure}

\subsection{RQ3: How are different experts in MoE uniquely aligned with different personalized features?}\label{sec:rq3}
% To understand whether different experts encode distinct personalized features during MoE LLMs' personalization, we evaluate how uniquely these experts align with different latents. Our \textbf{RQ2} (\S\ref{sec:rq2}) addresses the unique alignment between latents and personalized features and verifies it. To verify the unique alignment between experts and personalized features, we evaluate the alignment uniqueness between latents and experts, as it can be mapped via \textbf{RQ2}. We select the top-$N$ activating examples for each latent $j$ and analyze the distinctiveness and stability of top-$k'$ experts from a grid search over $(N,k)$ optimizing the two objectives (refer \Cref{sec:crosscoder_training_method}): \textit{Distinctiveness)} ($D$) and \textit{Stability} ($S$), where $N$ is the number of all examples activated by the latent $j$ and $k$ is the number of experts activated by the sparse routing mechanism for each token in MoE LLMs.
To test whether distinct personalization features map to different experts, we evaluate the distinctiveness and stability of latent–expert mappings (see \S\ref{sec:crosscoder_training_method}). As illustrated in \Cref{fig:latent_to_expert_mapping_hyperparameter_qwen}, for each latent, we select the optimal 400 examples and restrict it to route towards one expert every layer for the Qwen3-30B-A3B (2507). For this setting, we achieve very high stability with $S$ of 0.81, verifying that under repeated sampling over those 400 examples, most of the activating examples are routed towards a specialized expert. To validate whether the routing is heavily influenced by a dominant expert, we measure the distinctiveness and a very high $D$ of 0.94, which verifies that a different set of examples activated by different latents is routed to their respective experts. \textit{Overall, experts in MoE LLMs encode distinct personalized features, verifying lower polysemanticity for personalization.}\looseness=-1

\subsection{Causal analysis}\label{sec:causal_analysis}
% Our causal analysis verifies the (1) sufficiency of personalized latents in explaining emerging personalized features encoded via experts in MoE LLMs during personalization, and (2) whether specific experts encode distinct personalized features, indicating unique routing paths for personalization.  
\textbf{Sufficiency of personalized latents via activation patching.}
Following \citet{minder2025overcoming} of causally verifying the sufficiency of chat-specific latents for instruction-tuned LLMs, for an input token, we patch the activation of that token at the crosscoder layer of the personalized model with a hybrid activation combining activation from the non-personalized model and personalized latents activations for that token, and continue the generation from the later layer to measure the KL divergence between the next token probability of the personalized and patched personalized model. In particular, for an input token $x_t$ with activations from personalized and non-personalized MoE LLMs at the crosscoder layer $l$ as $h_{x_t}^{l, \text{ft}}$, and $h_{x_t}^{l, \text{base}}$ respectively, we measure KL divergence $\mathrm{KL}\left( p^{\mathrm{ft}}_{h_{x_t}^{l,\mathrm{ft}} \leftarrow h_{x_t}^{l,\mathcal{J}_{\mathrm{ft}}}} \,\middle\|\, p^{\mathrm{ft}} \right)$, where $p^{\text{ft}}$ is the personalized model's next token probability, and the hybrid activation $h_{x_t}^{l,\mathcal{J}_{\mathrm{ft}}} = h_{x_t}^{l,\mathrm{base}} + \sum_{j \in \mathcal{J}_{\mathrm{ft}}} a_j(x_t)\, \mathbf{d}_j^{\mathrm{ft}}$. If we just patch with the non-personalized model, then the KL divergence will be the maximum, and we measure the reduction in KL divergence from the non-personalized patching for our personalized latents. We also randomly select the same number of latents matching the activation frequency of those personalized latents, from the whole and `shared' region of the crosscoder's latent distribution as baselines. The top part of \Cref{fig:kl_patching_latent_expert_boosting} verifies that for personalized latents, the KL divergence decrease is the maximum among the baselines across MoE models on the MULTIFACETED BENCH benchmark, \textit{causally establishing the sufficiency of personalized latents in explaining personalized features in MoE personalization.}

\textbf{Unique routing paths for distinct personalization features.}
To causally verify whether a small set of experts encodes distinct personalization features as shown in \S\ref{sec:rq3}, we steer those experts in the non-personalized MoE LLMs along each dimension from the MULTIFACETED COLLECTION benchmark, and measure the KL divergence from the personalized counterparts. In particular, for each dimension, we collect experts uniquely aligned with the latents (\S\ref{sec:rq2},\S\ref{sec:rq2}) and rank them by routing frequency (\Cref{eq:exp_freq}) over the activating examples corresponding to those latents. We steer top-$k'$ (\Cref{fig:latent_to_expert_mapping_hyperparameter_omloe,fig:latent_to_expert_mapping_hyperparameter_qwen,fig:latent_to_expert_mapping_hyperparameter_mixtral}) experts for each dimension for all the MoE LLMs as: $\tilde{z}^{(l)}_{i} \;=\; z^{(l)}_{i} + \beta^{(l)} \sigma(z^{(l)}_{i})$, where $i \in \text{top}-k'$, $\sigma(z^{(l)}_{i})$ is the standard deviation of the router logits of the non-personalized MoE at layer $l$ (over tokens and experts) and $\beta^{(l)}$ (2 in our experiment) is the boosting strength. We also randomly steer the same number of experts at the crosscoder layer in the non-personalized layer. The bottom panel of \cref{fig:kl_patching_latent_expert_boosting} shows that, for each dimension, steering the unique top-$k'$ experts yields lower KL divergence than random steering, thereby \textit{causally verifying unique routing paths for distinct personalization features.}\looseness=-1

\section{Discussion and Conclusion}
We presented, to our knowledge, the first mechanistic study of personalization in MoE LLMs. Combining a two-stage router-and-expert adaptation for personalization with crosscoder-based interpretability analysis for personalization mechanism shows: (1) MoE LLMs achieve competitive personalization performance than dense LLMs by utilizing only $\sim20\%$ of total parameters. (2) MoE LLMs exhibit lower polysemanticity than dense LLMs for personalization and they enable better interpretability of encoded personalization features via crosscoder than dense LLMs. (3) MoE encodes distinct personalization features via different routing paths, enabling better localization and subsequent intervention for controlled personalization.  

% \clearpage

\section*{Limitations}

\textbf{Models and comparisons.}
We study three open MoE families and analyze one layer per model, selected by routing divergence. Personalization features may also arise at other layers, and our conclusions may not transfer to other MoE designs (e.g., shared experts or fine-grained expert segmentation). The dense counterparts are matched by active or total parameter count, but differ in pre-training data and
recipes, so architectural effects cannot be fully separated from training differences.

\textbf{Data and notion of persona.}
Multifaceted-Collection describes preferences through synthetic system messages over four dimensions, with categorical, long-tailed subdimension labels. This captures group-level preferences rather than individual users, and our experiments are limited to English. Personalization signals in real user histories may be noisier and more entangled than in this controlled setting.

\textbf{Crosscoder fidelity.}
Crosscoders reconstruct activations imperfectly, and reconstruction quality varies across models (with fractions of variance explained of $0.80$ for OLMoE, $0.74$ for Qwen3-30B-A3B, and $0.65$ for Mixtral, whose activations have much larger magnitudes and heavier tails). Information in the reconstruction error is invisible to our latent-level analyses, and the set of personalized latents depends on the decoder-norm threshold. BatchTopK training and latent scaling mitigate known sparsity artifacts \citep{minder2025overcoming}, but do not rule them out entirely.

\textbf{Probing.}
Persona concepts are properties of whole responses, whereas sparse probing labels individual tokens, thereby introducing label noise and lowering absolute scores across all models. Selecting the best of many components favors MoE models, which have more candidates (layers $\times$ experts); we mitigate this by selecting on test data and reporting test scores, but selection effects may remain, particularly for $k=1$. Probes measure whether a concept is decodable, not whether the model uses it.

\textbf{Expert attribution and causality.}
Expert sets are derived from routing frequencies at firing tokens and are therefore correlational; they depend on the number of examples, the selection rule, and the background routing rate, and their stability is moderate.

\textbf{Evaluation of generation.}
ROUGE-L and METEOR measure overlap with reference responses and are imperfect proxies for personalization: a response can satisfy a preference while diverging from the reference wording, as our steering results illustrate. Sampling-based generation adds variance. Automated explanations rely on an LLM explainer and scorer (Llama-3.3-70B-Instruct), whose own biases may shape the
descriptions.

\section*{Ethical Considerations}
\noindent \textbf{Use of AI Assistants.} The authors used Claude API to assist with language polishing and grammatical corrections, clarity and coherence. All technical contributions, including experimental design choices and final decisions, were made by the authors.

\section*{Acknowledgements}
$\begin{array}{l}\includegraphics[width=1cm]{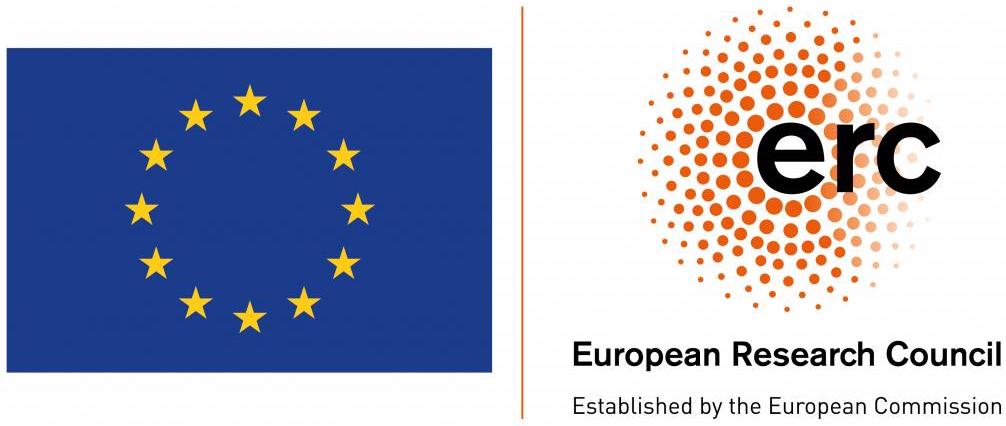} \end{array}$ 
This research was co-funded by the European Union (ERC, ExplainYourself, 101077481) and by VILLUM FONDEN (grant number 40543). Views and opinions expressed are however those of the author(s) only and do not necessarily reflect those of the European Union or the European Research Council. Neither the European Union nor the granting authority can be held responsible for them.

\bibliography{custom}

\appendix

\section{Relevant Model Architecture Details}\label{app:archi_details}
\subsection{Crosscoder}\label{sec:crosscoders_detailed}
Crosscoder is an encoder-decoder-based architecture that explains the activations from both finetuned (personalized) and base (non-personalized) LLMs by learning a shared dictionary of interpretable concepts (latents). Formally, for the input token $x_t$, let the activations from a particular layer $l$ of the personalized and non-personalized MoEs be $h_{x_t}^{(l,\text{ft})} \in \mathbb{R}^d$ and $h_{x_t}^{(l,\text{base})} \in \mathbb{R}^d$ respectively ($d$ is the model dimension), the encoder of the crosscoder maps those activations to a higher dimensional ($D>>d$)\footnote{$D=\alpha \times d$, where $\alpha$ is the expansion factor, and we keep it as 32 for our experiments.} latent space, where each dimension $j \in \{1, \ldots, D\}$ explains distinct features encoded in those activations $h_{x_t}^{(l,\text{ft})}$, and $h_{x_t}^{(l,\text{base})}$ as:
\begin{equation}
    a_j (x_t) = \text{ReLU}(\mathbf{e}_j^{(\text{ft})}h_{x_t}^{(l,\text{ft})} + \mathbf{e}_j^{(\text{base})}h_{x_t}^{(l,\text{base})} + \mathbf{b}_j^{(\text{enc})})
\end{equation}
where $a_j$ is the activation for latent $j$, $\mathbf{e}_j^{(\text{ft})}, \mathbf{e}_j^{(\text{base})} \in \mathbb{R}^d$ are the the encoder weight vectors corresponding to the latent $j$, and $\mathbf{b}_j^{(\text{enc})}$ is the corresponding encoder bias. The decoder of the crosscoder then reconstructs those activations $h_{x_t}^{(l,\text{ft})}$, and $h_{x_t}^{(l,\text{base})}$ for the token $x_t$ using the sparse activation distribution of shared dictionary latents as:
\begin{equation}
\begin{aligned}
    \tilde{h}_{x_t}^{(l,\text{ft})} &= \sum_j a_j(x_t)\, \mathbf{d}_j^{(\text{ft})} + \mathbf{b}^{(\text{dec})} \\
    \tilde{h}_{x_t}^{(l,\text{base})} &= \sum_j a_j(x_t)\, \mathbf{d}_j^{(\text{base})} + \mathbf{b}^{(\text{dec})}
\end{aligned}
\end{equation}
where $\mathbf{d}_j^{(\text{ft})}, \mathbf{d}_j^{(\text{base})} \in \mathbb{R}^d$ are the the decoder weight vectors corresponding to the latent $j$, and $\mathbf{b}_j^{(\text{dec})} \in \mathbb{R}^d$ is the decoder bias. The crosscoder is trained by minimizing the loss involving (a) reconstruction error as $h_{x_t}^{(l,\text{ft})} - \tilde{h}_{x_t}^{(l,\text{ft})}$, and $h_{x_t}^{(l,\text{base})} - \tilde{h}_{x_t}^{(l,\text{base})}$, and (b) sparsity enforcement on the decoder weights $\mathbf{d}_j^{(\text{ft})}, \mathbf{d}_j^{(\text{base})}$. Following \citep{minder2025overcoming} of utilizing L0 sparsity instead of L1 sparsity in obtaining better decoupled latents, we adopt the L0 sparsity-based BatchTopK crosscoder \citep{bussmann2024batchtopk} to select top-$k$ latents activated per example for a batch size of $n$, enforcing only $nk$ activated latents for activation reconstruction from both models. More details about BatchTopk crosscoder are described in \citet{minder2025overcoming}.

\section{Reproducibility}\label{app:reproduce}

\subsection{Performance comparison of dense and MoE LLMs for personalized text generation}

\begin{table}[t]
\centering
\small
\setlength{\tabcolsep}{4pt}
\renewcommand{\arraystretch}{1.15}
\begin{tabular}{lcc}
\toprule
\textbf{Model} & \textbf{Active} & \textbf{Total} \\
               & \textbf{Params} & \textbf{Params} \\
\midrule
\multicolumn{3}{l}{\textit{OLMoE family}} \\
\quad OLMo-2-1B (0425)       & 1B   & 1B   \\
\quad OLMoE-1B-7B (0125)     & 1.3B & 6.9B \\
\quad OLMo-2-7B (1124)       & 7B   & 7B   \\
\midrule
\multicolumn{3}{l}{\textit{Qwen family}} \\
\quad Qwen2.5-3B             & 3B   & 3B   \\
\quad Qwen3-30B-A3B (2507)   & 3B   & 30B  \\
\quad Qwen2.5-32B            & 32B  & 32B  \\
\midrule
\multicolumn{3}{l}{\textit{Mixtral / Llama-2 family}} \\
\quad Llama-2-13B            & 13B  & 13B  \\
\quad Mixtral-8x7B (v0.1)    & 13B  & 46.7B \\
\quad Llama-2-70B            & 70B  & 70B  \\
\bottomrule
\end{tabular}
\caption{For each MoE LLM, we compare with similar active and total parameter-sized dense LLMs with the same model family if available. For dense LLMs, active parameters are the same as total parameters.}
\label{tab:model_params}
\end{table}

\subsubsection{Hyperparameters}
Following \citep{lee2024aligning}, for inference on the multifacted-bench dataset, we utilize a temperature of $1.0$, a top-p of $0.9$, and a repetition penalty of $1.03$ for the generation configuration, with a maximum number of new tokens to generate of $4096$.

\subsubsection{Implementation Details}
We utilize LoRA-PEFT for both router tuning and personalized experts tuning of MoE models and personalization of dense LLMs. We only update the MLP weights for both MoE and dense LLMs. To train the crosscoder, we utilize the BatchTopK crosscoder with L0 and an expansion factor of 32 for all three MoE LLMs, using paired activations collected from the residual MLP contributions.

\begin{table*}[t]
\centering
\definecolor{denseA}{HTML}{E7F2F8}% Dense (active)  -- 25% tint of #9ECAE1
\definecolor{moe}{HTML}{FADDD1}%    MoE             -- 20% tint of #E6550D
\definecolor{denseT}{HTML}{B7D3E8}% Dense (total)   -- 35% tint of #3182BD
\resizebox{\textwidth}{!}{%
\setlength{\tabcolsep}{3.5pt}
\begin{tabular}{lc*{7}{cc}}
\toprule
\multirow{3}{*}{\textbf{Model}} & \multirow{3}{*}{\textbf{Act.}} & \multicolumn{2}{c}{\multirow{2}{*}{\textbf{MFBench}}} &
\multicolumn{6}{c}{\textbf{LaMP}} & \multicolumn{6}{c}{\textbf{LongLaMP}} \\
\cmidrule(lr){5-10} \cmidrule(lr){11-16}
 & & & & \multicolumn{2}{c}{\textbf{News}} & \multicolumn{2}{c}{\textbf{Scholar}} & \multicolumn{2}{c}{\textbf{Tweet}} &
\multicolumn{2}{c}{\textbf{Abstr.}} & \multicolumn{2}{c}{\textbf{Topic}} & \multicolumn{2}{c}{\textbf{Review}} \\
\cmidrule(lr){3-4} \cmidrule(lr){5-6} \cmidrule(lr){7-8} \cmidrule(lr){9-10} \cmidrule(lr){11-12} \cmidrule(lr){13-14} \cmidrule(lr){15-16}
 & & M & R & M & R & M & R & M & R & M & R & M & R & M & R \\
\midrule
\rowcolor{denseA}OLMo-2-1B (0425) & 1B & 0.29 & 0.19 & \underline{0.12} & 0.08 & 0.30 & 0.18 & 0.29 & 0.22 & \textbf{0.31} & 0.18 & \underline{0.23} & 0.11 & 0.18 & \underline{0.14} \\
\rowcolor{denseA}OLMo-2-1B (0425)$^{\star}$ & 1B & 0.31 & 0.20 & \underline{0.12} & \underline{0.12} & 0.38 & 0.29 & \underline{0.40} & \textbf{0.38} & 0.26 & 0.19 & \underline{0.23} & \underline{0.12} & 0.14 & 0.12 \\
\midrule
\rowcolor{moe}OLMoE-1B-7B (0125) & 1.3B & 0.25 & 0.20 & \underline{0.12} & 0.11 & 0.36 & 0.28 & 0.35 & 0.32 & \textbf{0.31} & 0.18 & \textbf{0.24} & 0.11 & \textbf{0.21} & \underline{0.14} \\
\rowcolor{moe}OLMoE-1B-7B (0125)$^{\dagger}$ & 1.3B & \underline{0.32} & 0.20 & \underline{0.12} & \underline{0.12} & 0.35 & 0.26 & 0.36 & 0.32 & \textbf{0.31} & 0.17 & \underline{0.23} & 0.11 & \underline{0.19} & \underline{0.14} \\
\rowcolor{moe}OLMoE-1B-7B (0125)$^{\ddagger}$ & 1.3B & \underline{0.32} & 0.21 & \underline{0.12} & \underline{0.12} & 0.37 & 0.29 & \underline{0.40} & \underline{0.37} & \underline{0.30} & 0.18 & \underline{0.23} & \underline{0.12} & 0.14 & 0.13 \\
\midrule
\rowcolor{denseT}OLMo-2-7B (1124) & 7B & 0.30 & \underline{0.23} & \textbf{0.13} & \textbf{0.13} & \textbf{0.40} & \textbf{0.32} & 0.36 & 0.33 & \underline{0.30} & \underline{0.20} & \textbf{0.24} & \underline{0.12} & \textbf{0.21} & \textbf{0.15} \\
\rowcolor{denseT}OLMo-2-7B (1124)$^{\star}$ & 7B & \textbf{0.34} & \textbf{0.24} & \textbf{0.13} & \textbf{0.13} & \underline{0.39} & \underline{0.30} & \textbf{0.41} & \textbf{0.38} & \underline{0.30} & \textbf{0.21} & 0.21 & \textbf{0.13} & 0.16 & \underline{0.14} \\
\midrule \midrule
\rowcolor{denseA}Qwen2.5-3B & 3B & 0.31 & 0.23 & \underline{0.13} & 0.11 & 0.37 & 0.27 & \underline{0.40} & 0.36 & \textbf{0.34} & 0.20 & \underline{0.24} & \underline{0.13} & \underline{0.24} & \textbf{0.16} \\
\rowcolor{denseA}Qwen2.5-3B$^{\star}$ & 3B & \underline{0.35} & 0.24 & 0.12 & 0.12 & 0.39 & 0.31 & 0.38 & 0.35 & \underline{0.32} & 0.20 & 0.20 & 0.12 & 0.21 & \underline{0.15} \\
\midrule
\rowcolor{moe}Qwen3-30B-A3B (2507) & 3B & 0.31 & 0.24 & \textbf{0.14} & \textbf{0.14} & 0.32 & 0.23 & \textbf{0.41} & \textbf{0.40} & \underline{0.32} & \underline{0.21} & 0.22 & \underline{0.13} & \underline{0.24} & \underline{0.15} \\
\rowcolor{moe}Qwen3-30B-A3B (2507)$^{\dagger}$ & 3B & \underline{0.35} & \underline{0.25} & \textbf{0.14} & \textbf{0.14} & \textbf{0.42} & \textbf{0.35} & \textbf{0.41} & \textbf{0.40} & 0.31 & \underline{0.21} & 0.22 & \underline{0.13} & 0.23 & \textbf{0.16} \\
\rowcolor{moe}Qwen3-30B-A3B (2507)$^{\ddagger}$ & 3B & \textbf{0.37} & \textbf{0.26} & \underline{0.13} & \underline{0.13} & \underline{0.41} & \underline{0.33} & 0.39 & \underline{0.37} & \underline{0.32} & \underline{0.21} & 0.22 & \underline{0.13} & 0.23 & \textbf{0.16} \\
\midrule
\rowcolor{denseT}Qwen2.5-32B & 32B & 0.33 & \textbf{0.26} & \textbf{0.14} & 0.12 & 0.33 & 0.18 & 0.39 & 0.35 & 0.31 & \textbf{0.23} & \textbf{0.25} & \textbf{0.14} & \underline{0.24} & \textbf{0.16} \\
\rowcolor{denseT}Qwen2.5-32B$^{\star}$ & 32B & \textbf{0.37} & \textbf{0.26} & \underline{0.13} & \underline{0.13} & \underline{0.41} & \underline{0.33} & 0.37 & 0.33 & \underline{0.32} & \underline{0.21} & 0.23 & 0.12 & \textbf{0.25} & \underline{0.15} \\
\midrule \midrule
\rowcolor{denseA}Llama-2-13b & 13B & 0.27 & 0.20 & \underline{0.13} & 0.07 & 0.29 & 0.14 & 0.33 & 0.26 & \textbf{0.33} & 0.21 & \underline{0.23} & \underline{0.12} & \underline{0.23} & \underline{0.15} \\
\rowcolor{denseA}Llama-2-13b$^{\star}$ & 13B & \underline{0.34} & \underline{0.23} & \underline{0.13} & \textbf{0.13} & 0.40 & 0.30 & \textbf{0.38} & \textbf{0.36} & \underline{0.32} & 0.19 & \underline{0.23} & \underline{0.12} & 0.19 & 0.14 \\
\midrule
\rowcolor{moe}Mixtral-8x7B (v0.1) & 13B & 0.27 & 0.22 & \textbf{0.14} & \textbf{0.13} & 0.41 & 0.29 & 0.30 & 0.23 & 0.31 & \textbf{0.24} & \underline{0.23} & \textbf{0.13} & 0.22 & \textbf{0.16} \\
\rowcolor{moe}Mixtral-8x7B (v0.1)$^{\dagger}$ & 13B & 0.30 & 0.22 & \underline{0.13} & \textbf{0.13} & \textbf{0.43} & \textbf{0.34} & 0.34 & 0.33 & \underline{0.32} & \textbf{0.24} & 0.22 & \textbf{0.13} & 0.18 & 0.14 \\
\rowcolor{moe}Mixtral-8x7B (v0.1)$^{\ddagger}$ & 13B & \textbf{0.36} & \textbf{0.25} & \underline{0.13} & \textbf{0.13} & \underline{0.42} & \underline{0.32} & \textbf{0.38} & \textbf{0.36} & 0.31 & \underline{0.22} & \textbf{0.24} & \textbf{0.13} & 0.22 & \underline{0.15} \\
\midrule
\rowcolor{denseT}Llama-2-70b & 70B & 0.28 & 0.22 & \textbf{0.14} & \underline{0.10} & 0.27 & 0.13 & 0.30 & 0.26 & \underline{0.32} & 0.21 & \textbf{0.24} & \textbf{0.13} & \textbf{0.24} & \underline{0.15} \\
\rowcolor{denseT}Llama-2-70b$^{\star}$ & 70B & \textbf{0.36} & \textbf{0.25} & \textbf{0.14} & \textbf{0.13} & 0.41 & \underline{0.32} & \underline{0.36} & \underline{0.34} & 0.30 & 0.21 & \textbf{0.24} & \textbf{0.13} & \underline{0.23} & \underline{0.15} \\
\midrule \midrule
JANUS 7B & 7B & 0.35 & 0.24 & 0.13 & 0.13 & 0.40 & 0.32 & 0.37 & 0.35 & 0.30 & 0.21 & 0.24 & 0.13 & 0.23 & 0.15 \\
JANUS+ORPO 7B & 7B & 0.21 & 0.10 & 0.12 & 0.05 & 0.09 & 0.05 & 0.29 & 0.15 & 0.15 & 0.09 & 0.20 & 0.10 & 0.07 & 0.04 \\
JANUS+DPO 7B & 7B & 0.35 & 0.23 & 0.14 & 0.13 & 0.41 & 0.32 & 0.36 & 0.34 & 0.31 & 0.22 & 0.24 & 0.13 & 0.24 & 0.15 \\
\bottomrule
\end{tabular}%
}
\caption{Personalization performance comparison over LLMs across personalization benchmarks using two evaluation metrics METEOR (M) and ROUGE (R). Act.\ indicates active parameters of LLMs. $^{\star}$ indicates PEFT fine-tuned (FT) dense LLMs, $^{\dagger}$ indicates router-tuned (RT) MoE LLMs, and $^{\ddagger}$ indicates the two stage router and expert (RT + EFT)-tuned MoE LLMs. Row colors: \colorbox{denseA}{dense, active size}, \colorbox{moe}{MoE}, \colorbox{denseT}{dense, total size}. Acronym of personalization benchmarks: \textbf{MFBench} = MULTIFACETED BENCH; \textbf{LaMP}: News = News Headline Generation, Scholar = Scholarly Title Generation, Tweet = Tweet Paraphrasing; \textbf{LongLaMP}: Abstr.\ = Abstract Generation, Topic = Topic Writing, Review = Review Generation. Within each model family (dense and MoE models together), metric and benchmark, \textbf{bold} indicates the best and \underline{underline} indicates the second-best result, and tied values share the mark.}
\label{tab:llm_personalization_comp}
\end{table*}

% \section{Additional Results}\label{app:additional_results}

\begin{figure}[t]
  \centering
  \includegraphics[width=0.9\columnwidth]{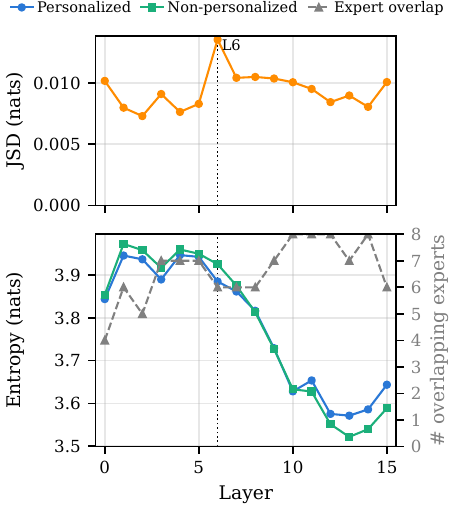}
  \caption{Layer-wise routing distribution analysis. \textbf{Top:} routing JS-divergence between the personalized and non-personalized OLMoE-1B-7B (0125. \textbf{Bottom:} entropy of the two models with overlapped experts (right axis). L6 is identified as the best layer.}
  \label{fig:layerwise_routing_jsd_entropy_omloe}
\end{figure}

\begin{figure}[t]
  \centering
  \includegraphics[width=0.9\columnwidth]{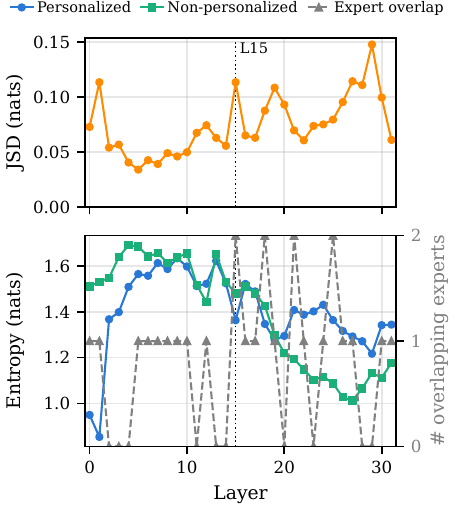}
  \caption{Layer-wise routing distribution analysis. \textbf{Top:} routing JS-divergence between the personalized and non-personalized Mixtral-8x7B (v0.1). \textbf{Bottom:} entropy of the two models with overlapped experts (right axis). L15 is identified as the best layer.}
  \label{fig:layerwise_routing_jsd_entropy_mixtral}
\end{figure}

\begin{figure}[t]
    \centering
    \includegraphics[width=0.9\linewidth]{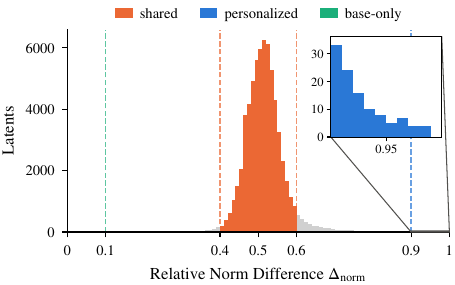}
    \caption{Relative norm difference distribution of latents from the crosscoder trained on activations from L6 of personalized and non-personalized OLMoE-1B-7B (0125).}
    \label{fig:latent_norm_dist_olmoe}
\end{figure}

\begin{figure}[t]
    \centering
    \includegraphics[width=0.9\linewidth]{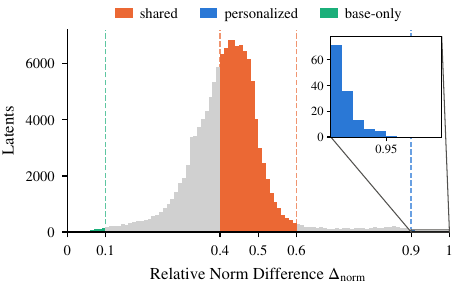}
    \caption{Relative norm difference distribution of latents from the crosscoder trained on activations from L15 of personalized and non-personalized Mixtral-8x7B (v0.1).}
    \label{fig:latent_norm_dist_mixtral}
\end{figure}

\begin{table*}[t]
\centering
\small
\setlength{\tabcolsep}{3pt}
\renewcommand{\arraystretch}{1.4}
\begin{tabular}{p{2.2cm} c p{3.5cm} p{2.4cm} p{1.8cm} p{3.5cm}}
\toprule
\textbf{Model} & \textbf{Latent} & \textbf{Explanation} & \textbf{Dimension} & \textbf{Top-$k'$ Experts} & \textbf{Example} \\
\midrule
OLMoE-1B-7B (0125)
  & 12503
  & The feature detects the presence of meticulously structured and precise language, often characterized by the use of advanced technical terminology, detailed explanations, and a formal tone, typically in the context of scientific, technical, or academic discussions.
  & informativeness
  & L6-E9 (0.91);L6-E22 (0.73)
  & \textit{``\ldots Your discussions are \textcolor{red}{\underline{met}iculously} detailed, exploring biochemical mechanisms of Vitamin C\ldots''} \\
\addlinespace[3pt]
Mixtral-8x7B (v0.1)
  & 45358
  & The feature detects tokens related to grammatical correctness, punctuation, and sentence structure, often in the context of editing, correcting, or explaining language usage.
  & style
  & L15-E2 (0.96);L15-E5 (0.34)
  & \textit{``\ldots identifying run-on sentences with a focus on \textcolor{red}{gram\underline{mat}ical} terms, catering to an audience with prior knowledge\ldots''} \\
\bottomrule
\end{tabular}
\caption{Examples of personalized latents identified by the crosscoder for three MoE LLMs: autointerpretability explanation, dominant preference dimension from the MULTIFACETED COLLECTION benchmark, top-$k'$ experts at the optimal layer with the percentage of activated token routed towards that expert, and a representative example from top-$N$ maximally activating sequences with the maximally activated token highlighted.}
\label{tab:latent_analysis}
\end{table*}

\begin{figure}[t]
  \centering
  \includegraphics[width=0.9\columnwidth]{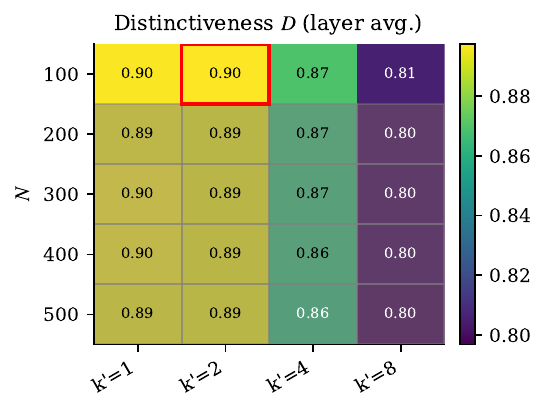}\\
  \includegraphics[width=0.9\columnwidth]{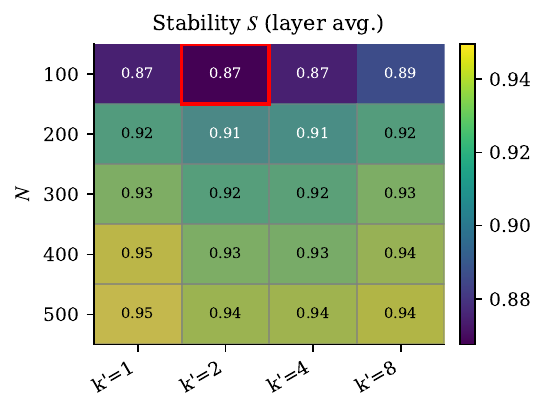}
  \caption{Grid search over $(N,k)$ for OLMoE-1B-7B (0125). \textbf{Top:} Distinctiveness of top-$k'$ experts across latents with varying $N$. \textbf{Bottom:} Stability of top-$k'$ experts for the same $N$ examples under resampling for a particular latent.}
  \label{fig:latent_to_expert_mapping_hyperparameter_omloe}
\end{figure}

\begin{figure}[t]
  \centering
  \includegraphics[width=0.9\columnwidth]{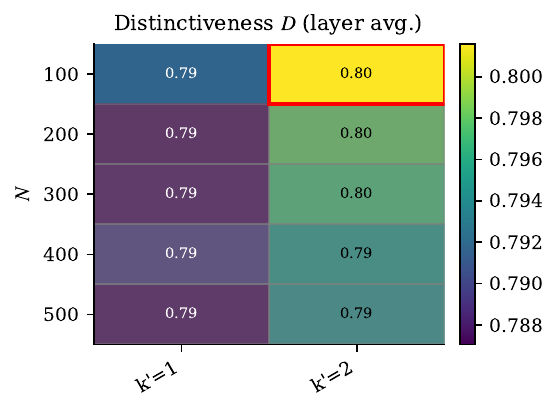}\\
  \includegraphics[width=0.9\columnwidth]{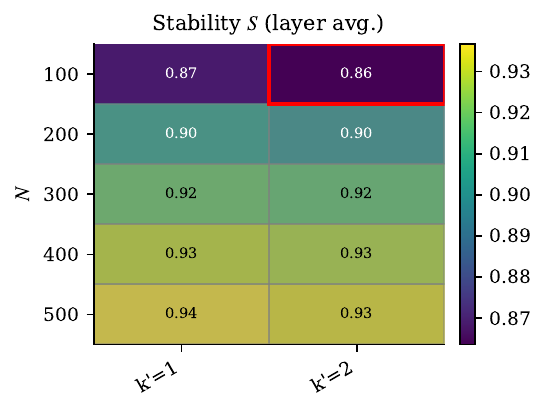}
  \caption{Grid search over $(N,k)$ for Mixtral-8x7B (v0.1). \textbf{Top:} Distinctiveness of top-$k'$ experts across latents with varying $N$. \textbf{Bottom:} Stability of top-$k'$ experts for the same $N$ examples under resampling for a particular latent.}
  \label{fig:latent_to_expert_mapping_hyperparameter_mixtral}
\end{figure}

\begin{figure}[t]
  \centering
  \includegraphics[width=0.9\columnwidth]{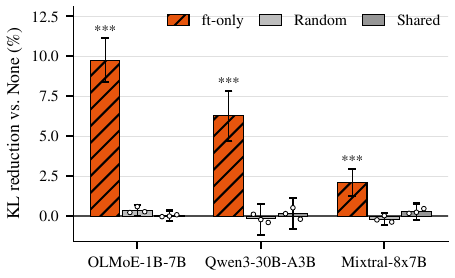}
  \caption{KL patching at the crosscoder layer over the first 9 response tokens. Reduction of the next-token KL divergence to the personalized MoE, relative to patching in the non-personalized MoE output, when adding the decoder contributions of the personalized latents (ft-only) or of equally sized, firing-frequency-matched random or shared latents (mean of 3 draws; dots: individual draws). Error bars: 95\% bootstrap CIs over the 921 \textsc{Multifaceted Bench} examples; ***: $p<0.001$ vs.\ both controls (Wilcoxon, Holm-corrected).}
  \label{fig:kl_patching}
\end{figure}

\end{document}